\newif\ifarxiv
\arxivtrue 

\ifarxiv
\documentclass[twocolumn]{article}
\newcommand{\IEEEmembership}[1]{}
\newenvironment{IEEEkeywords}{%
  \bfseries\textit{Index Terms}---%
}{%
}
\newenvironment{IEEEbiography}[2][]{}{}

\usepackage[letterpaper, top=0.6in, bottom=0.5in, inner=0.75in, outer=0.75in, headheight=0.3in, footskip=0.3in]{geometry}
\usepackage{fancyhdr}
\usepackage{times}

\else
    \documentclass[journal]{IEEEtran}
\fi

\usepackage[pdftex]{graphicx}
\usepackage[cmex10]{amsmath}
\usepackage{algorithmic}
\usepackage{xcolor}
\usepackage{array}
\usepackage{stfloats}
\usepackage{hyperref}
\usepackage{url}
\usepackage{verbatim}
\usepackage[sort&compress, numbers]{natbib}
\usepackage{caption}
\usepackage[T1]{fontenc}
\usepackage[utf8]{inputenc}
\usepackage{dirtytalk}

\begin{document}
%
\title{Terrain-Informed Self-Supervised Learning: Enhancing Building Footprint Extraction from LiDAR Data with Limited Annotations}
%
%
%

\ifarxiv
    \author{Anuja~Vats, David~Völgyes, Martijn~Vermeer, Marius~Pedersen,\\ Kiran~Raja,
            Daniele~S.~M.~Fantin and Jacob~Alexander~Hay%
    \thanks{Anuja Vats, Marius Pedersen and Kiran Raja are with the Department %
    of Computer Science, Norwegian Institute of Science and Technology, Gjøvik, Norway (e-mail: anuja.vats@ntnu.no, marius.pedersen@ntnu.no, kiran.raja@ntnu.no).}
    \thanks{David Völgyes, Martijn Vermeer, Daniele S. M. Fantin and Jacob Alexander Hay are with Science and Technology AS, Tordenskioldsgate 2, 0160 Oslo, Norway (e-mail: volgyes@stcorp.no, vermeer@stcorp.no, fantin@stcorp.no, hay@stcorp.no )}
    }
\fi

%
%

\markboth{Journal of \LaTeX\ Class Files,~Vol.~13, No.~9, September~2014}%
{Vats \MakeLowercase{\textit{et al.}}: Bare Demo of IEEEtran.cls for Journals}
%



\maketitle

\begin{abstract}
Estimating building footprint maps from geospatial data is vital in urban planning, development, disaster management, and various other applications. Deep learning methodologies have gained prominence in building segmentation maps, offering the promise of precise footprint extraction without extensive post-processing. However, these methods face challenges in generalization and label efficiency, particularly in remote sensing, where obtaining accurate labels can be both expensive and time-consuming.
To address these challenges, we propose terrain-aware self-supervised learning, tailored to remote sensing, using digital elevation models from LIght Detection and Ranging (LiDAR) data. We propose to learn a model to differentiate between bare Earth and superimposed structures enabling the network to implicitly learn domain-relevant features without the need for extensive pixel-level annotations.
We test the effectiveness of our approach by evaluating building segmentation performance on test datasets with varying label fractions. Remarkably, with only 1\% of the labels (equivalent to 25 labeled examples), our method improves over ImageNet pretraining, showing the advantage of leveraging unlabeled data for feature extraction in the domain of remote sensing. The performance improvement is more pronounced in few-shot scenarios and gradually closes the gap with ImageNet pretraining as the label fraction increases.
We test on a dataset characterized by substantial distribution shifts (including resolution variation and labeling errors) to demonstrate the generalizability of our approach. When compared to other baselines, including ImageNet pretraining and more complex architectures, our approach consistently performs better, demonstrating the efficiency and effectiveness of self-supervised terrain-aware feature learning.

\end{abstract}


%

\section{Introduction}
\label{sec:intro}
Estimating building footprint maps from geospatial data has been seen as indispensable for activities pertaining to urban planning and development such as estimating population distribution, monitoring urban expansion and impervious areas, creating detailed 3D city models, transportation, and navigation, as well as detecting illegal construction cases. In scenarios where timely information is crucial, such as assessing damage in the aftermath of natural disasters or updating topographical databases at a national scale, methods for creating accurate and updated maps of dynamic geographies are paramount.

Recent methods for building segmentation have shifted towards the application of deep learning methodologies~\cite{zhu2017deep}, offering the prospect of more precise building footprint extraction while minimizing the need for extensive post-processing. The segmentation task is difficult due to the vast heterogeneity across urban scenes globally. Factors such as variabilities across sensors, data quality, resolution, atmospheric and topographic differences, seasonal variations, etc. have been very challenging to counter when generalizing to unseen geographies~\cite{maggiori2017can}.
Further, the most noteworthy results in supervised learning in remote sensing stem from learning on a large number of accurately labeled images. Although open mapping platforms such as OpenStreetMap (OSM) can be used for acquiring labels, they may not be ideal due to inconsistencies between the vector data from the platform and the acquired data. Misalignments may arise due to differences in spatial resolution or time differences resulting in topography changes. The accuracy of labels derived from resolving for resolution differences can result in errors (such as those highlighted in later sections).  Moreover, as open platforms rely on contributions from individuals on a voluntary basis, the label density varies around the globe, where certain high-resource regions such as Europe have more labels than regions such as Africa or Asia \cite{janik2022sampling}. This may result in incomplete or outdated data.

For certain specialized remote sensing applications, the pre-defined categories in platforms like OSM may even be too general for use, requiring experts to annotate the data effectively. This is not only expensive but typically requires manual work by a domain expert~\cite{uzkent2019learning, chu2019geo}. In most real-world scenarios there is a shortage of labels at the desired level of preciseness. Segmentation tasks face a particular challenge exacerbated by the time-intensive nature of pixel-level labeling. Furthermore, in contrast to certain domains where data scarcity is a primary concern, remote sensing offers a unique advantage: extensive datasets collected over time, covering diverse geographic areas and conditions, sourced from satellites orbiting the Earth. This can be taken advantage of using alternative methods of acquiring supervision, particularly from unlabeled datasets. Recent studies have shown the value of leveraging the intrinsic structure and patterns inherent within unlabeled data. This learning paradigm, known as self-supervised learning derives supervisory signals from data through pretext tasks. Learning in this way enables discovering domain-relevant features that can greatly improve segmentation tasks of interest while requiring fewer pixel-level annotations~\cite{ayush2021self, dong2020self}.

This work targets the two aforementioned challenges that persist in the domain of remote sensing for improvement (i) generalization and (ii) label-efficiency. We propose a terrain-aware pretext task for learning features that can generalize despite commonly encountered domain shifts such as in input resolutions, geographies, and labeling errors (generalization) and perform well on downstream building segmentation with only a few labels (label efficiency).
Further, although the core motivation behind our design is to improve generalization performance while strictly constraining the annotation requirement, we also show that a focus on constructing a more efficient learning strategy allows simplifying computational complexity, where already available simple models like U-Net \cite{ronneberger2015unet} perform on par with computationally much more complex models such as transformers \cite{dosovitskiy2020image,aleissaee2023transformers} promoting efficiency and resource conservation. While the use of segmentation models based on complex architectures such as transformers has become mainstream \cite{aleissaee2023transformers}, in this paper, we argue in favour of investigating if the same advantages may be obtained at lower computation complexity, by improving learning strategies. To this end, we perform self-supervised learning instead of supervised learning, to learn generalizable features without labels. Since a self-supervised task such as ours is not learning via class-discrimination (as in many classification tasks), it can learn features that tend to be more informative of the input domain as opposed to discriminative. Such informative features lend to easy adaptation in the face of generalization, not just to out-of-distribution samples later on, but also to various tasks such as tree-species classification, land-use segmentation, etc.
The contributions of this work are:
\begin{itemize}
    \item We propose a new pretext task for learning generalizable features from unlabeled LiDAR data catering specifically to the domain of remote sensing considering the limited-label availability scenario.
    \item Our approach demonstrates enhanced performance compared to supervised ImageNet pretraining, resulting in a substantial reduction in the number of labels required for downstream training. Table \ref{tab:100_norway} displays the results in a few-shot learning setting, where we observe improved performance over ImageNet pretraining even with just 25 examples (equivalent to 1\% of the labels for fine-tuning). This indicates the efficacy of our terrain-aware pretext task in minimizing label requirements in the field of remote sensing.
    \item Our method also shows better generalization performance under radical domain shift between training and test datasets. We surpass the baseline model (U-Net Random) by a margin of 0.32 in Intersection over Union (IoU) (0.844 vs 0.521 Intersection over Union in Table \ref{tab:mapai} indicating the benefit from our learning strategy in feature reuse and transfer) and a transformer-based model by a margin of 0.054 in IoU  (0.844 vs 0.790 indicating gain in computational complexity).
\end{itemize}

\begin{figure}[!htbp]
    \centering
    \includegraphics[width=3.486in]{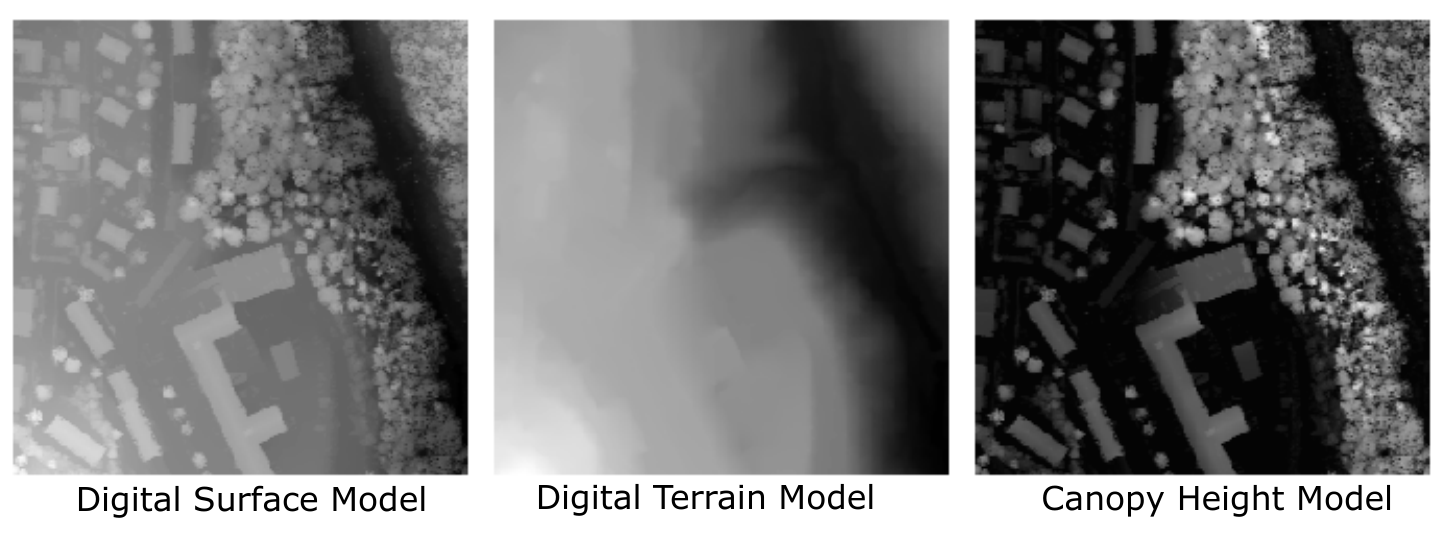}
    \caption{Different digital elevation models characterizing different aspects of the Earth's surface derived from LiDAR point clouds.}
    \label{fig:models}
\end{figure}

In this study, our focus shifts away from spectral data, and instead, we use the Norwegian national elevation model derived from aerial LiDAR data\footnote{https://hoydedata.no/ \label{hoyde}}. Specifically, we employ three distinct Digital Elevation Models (DEMs). These DEMs capture and characterize Earth's terrain in a three-dimensional context. A Digital Surface Model (DSM), represents the uppermost surface of the Earth's landscape. It encompasses all objects and terrain features, including natural elements like trees, man-made structures such as buildings, and other terrain irregularities. Essentially, it is a representation of the Earth's surface, including both ground and above-ground elements. A Digital Terrain Model (DTM) on the other hand, specifically isolates the bare Earth's topography, excluding any above-ground objects such as buildings and vegetation. It provides a representation of the ground surface free from the influence of man-made features or vegetation. Thus serving as a baseline elevation model, aiding in the identification of landforms and contours. A third model, known as the normalized Digital Surface Model (nDSM) (alternatively referred to as Canopy Height Model (CHM) in forestry applications when dealing with the canopy of the forest) is derived by subtracting the DTM from the DSM. This subtraction isolates the height of structures above ground enabling knowledge of vertical structures and the density of objects within a given area. These models (DSM, DTM, nDSM) are gridded (alternatively referred to as rasterizing), georeferenced, and 2D projected into local geospatial coordinate reference systems. The 3D LiDAR point cloud can be processed into DTM and DSM with traditional methods~\cite{Hug2004AdvancedLD, Li2016AGF}. Figure \ref{fig:models} shows the three different elevation datasets derived from the LiDAR data.

\section{Related work}
\subsection*{Building segmentation} Owing to its importance in urban planning, several approaches have been developed~\cite{vakalopoulou2015building,8710471,8711930,7552806,sym11010003} with the aim of continuously improving the detection of buildings across different landscapes. The authors in~\cite{vakalopoulou2015building, 7552806} use high-resolution images for improving building delineation. To overcome challenges associated with building displacements and shadows, integrating data from multiple sources, such as LiDAR in addition to spectral information from aerial and satellite images has been shown to enhance the robustness and accuracy of building segmentation~\cite{sohn2007data,8710471,hermosilla2011evaluation,ji2018fully}. Some approaches have also utilized stereo images \cite{yu2021automatic} to create stereo point clouds which in turn can be used to create surface maps. However, despite advantages~\cite{alonso2010satellite}, relatively fewer approaches depend solely on LiDAR data for segmentation. This is due to difficulties associated primarily with identifying and eliminating vegetation from LiDAR data as well as constraints related to cost and availability. Despite this, it is valuable to exploit the rich 3D structural information available from LiDAR data for building segmentation, independently from spectral information as fusing sources not only leaves room for introducing registration and resampling errors but allows developing potentially unnecessary reliance on an additional source during inference. In this work, we show that gridded LiDAR elevation data solely, without requiring elaborate pre-processing techniques, can lead to comparable or better performance on building segmentation as compared to methods that combine LiDAR data with alternative sensor information.

\subsection*{Self-supervised learning} Self-supervised learning (SSL) is learning by encoding unlabeled data to a lower-dimensional representation while being constrained to preserve information beneficial for downstream tasks such as classification. This step is often known as self-supervised pretraining. The supervisory signal during pretraining is derived from a surrogate/pretext task designed to extract desirable information (desirable to downstream tasks, while ignoring other task-irrelevant features) implicit within the data. A common approach is to use the network to predict the structural or semantic properties of the data. By learning to recover this information, features regarding color, texture, scale, etc. may be extracted. The encoded features are then transferred to (downstream) tasks of interest such as building or road segmentation, land use classification, etc. Learning in this way has been shown to perform better than supervised pretraining on many problems~\cite{goyal2021self, he2020momentum, chen2020improved}. In remote sensing, Zhang \textit{et al.}~\cite{zhang2019rotation} perform self-supervised rotation-angle prediction along with target recognition within a multi-task learning framework, where the bottom layers share parameters between the two tasks. Similarly, Zhao \textit{et al}.~\cite{zhao2020self} combine rotation prediction with scene classification, such that weight parameters for each loss are sampled randomly from a beta distribution. Tao \textit{et al.}~\cite{tao2020remote} analyze the impact of learning via different pretext tasks, these tasks being image inpainting, predicting relative position between image patches, and instance discrimination. Vincenzi et al.~\cite{vincenzi2021color} employ a colorization task for self-supervision, where the networks predict RGB channel information from other spectral bands. They emphasize the importance of tailoring pretext tasks specifically to remote sensing rather than translating directly from other domains to extract beneficial task-specific features. Further, typical self-supervision approaches like colorization, seasonal contrast or rotation prediction are not designed for remote sensing and thus are not applicable directly to LiDAR data. For example, data from different seasons may not be available for contrasting or rotated versions of LiDAR images may appear as natural as original images (as opposed to rotated image of a tree). A self-supervised task such as colorization or inpainting would necessitate an aligned secondary source additionally to LiDAR. Unlike them, we propose a  pretext task for elevation/LiDAR data catering specifically to remote sensing for learning effective and generalizable features from LiDAR data in a label-free manner.

\section{Methodology}
As discussed prior, self-supervision is an effective way to reduce the amount of training data required for a task by pretraining a model on a dummy pretext task \(T_{pretext}\), labeled data for which can be freely curated, \((X, Y_{pretext})\). The self-supervised learning objective can be defined as minimizing the discrepancy between predicted labels \(Y^{'}_{pretext}\) and artificially generated  \(Y_{pretext}\).
\begin{equation}
\label{eq:0}
    min_\theta = L_{pretext}(Y_{pretext},Y^{'}_{pretext})
\end{equation}

In this work, we propose a pretext task \(T_{pretext}\) to predict DTM (terrain) from DSM (surface). Essentially, the task is to remove all the above-ground structures such as vegetation, buildings, houses, powerlines, etc, to recover the underlying topography encompassing components of natural landscapes. However, since the effectiveness of this strategy depends on the extent to which the acquired representation from the pretext task \( \phi(X)\) can be transferred and minimally adapted for use in the downstream task, the choice of the task should be such as to encourage a unified and transferable feature space \(\phi(X)\) that can be effectively utilized for target task of interest.

Considering this requirement of pretext tasks, our \(T_{pretext}\) forces the network to differentiate between which elevation values correspond to bare-Earth as opposed to those of superimposed structures, such as vegetation and buildings. Thus encouraging the network to establish relationships between the various types of structures and the underlying terrain. The relationships could be the presence of fewer artificial objects on mountainous terrain compared to urban terrain, discernible variations in tree heights across diverse terrains, and fluctuations in building size, shape, and typology relative to the nature of the terrain. Our pretext task facilitates learning these relationships among different objects and their interplay with the terrain, without the need for explicit labels.

\begin{figure}[!ht]
   \centering
    \includegraphics[width=3.486in, height=4cm]{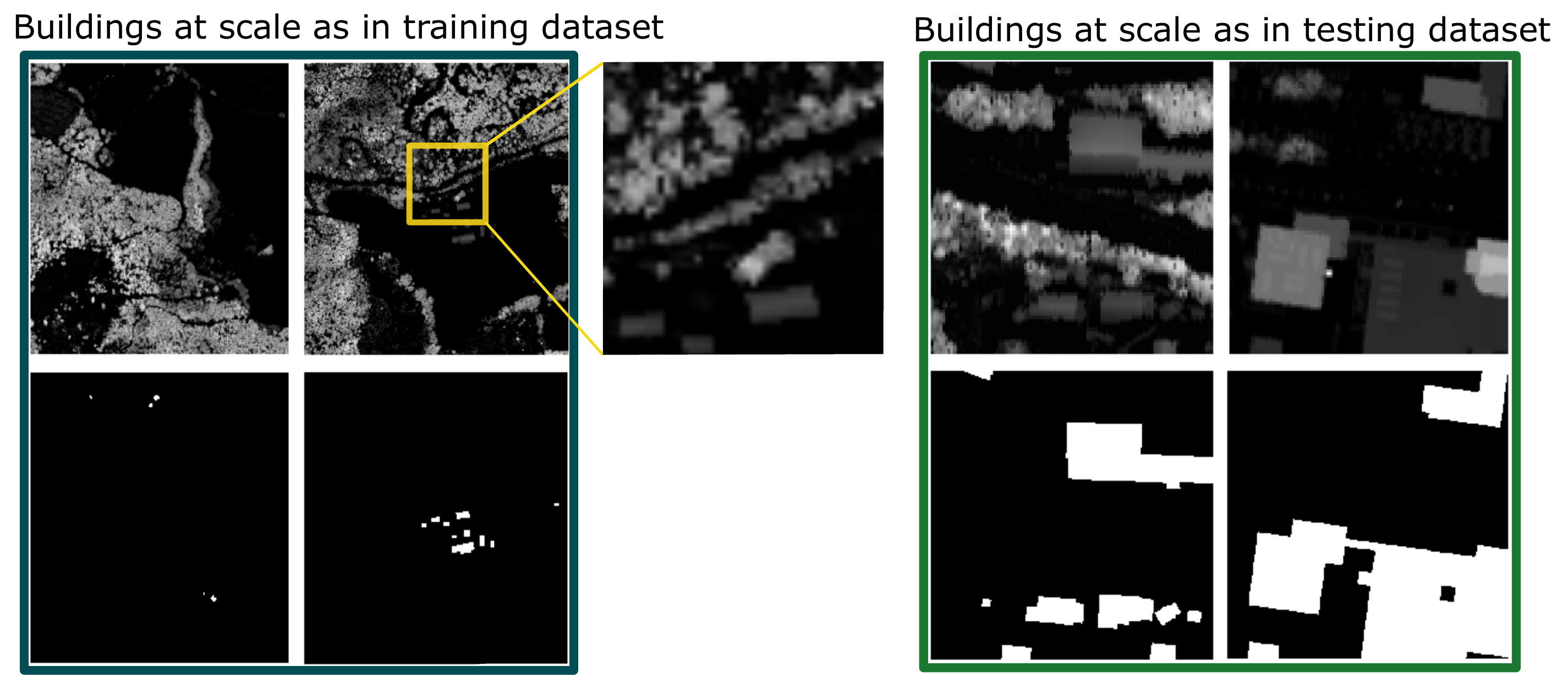}
    \caption{Scale differences as seen in our datasets. The top row shows LiDAR images at different scales from the train (left) and test (right) datasets. The corresponding building labels are shown in the bottom row. Even though our approach is pretrained for terrain-recovery on images as seen on the left, it translates easily to building with scales and sizes as seen on the right, indicating generalizability to objects at different scales (The magnification is to enable visual comparison between the two scales and aid reader comprehension and is not drawn to scale).}
    \label{fig:scales}
\end{figure}

Further, considering that segmentation networks struggle when it comes to identifying buildings that exhibit diverse shapes and scales~\cite{long2015fully}, it is advantageous to have a \(T_{pretext}\) that counters this. Also, the quality of labels plays a role as the DTM/DSM used for deriving segmentation labels may be inaccurate, further affecting learning. Refer to Figure \ref{fig:scales} which visually demonstrates the size and scale differences that we work with in our datasets. While several approaches have been developed to mitigate resolution differences between training and testing scenarios ~\cite{maggiori2016convolutional,yuan2017learning,wu2018automatic}, a degree of invariance to scale is implicit in our \(T_{pretext}\). This is because the expected scale of objects may be inferred indirectly based on the characteristics of the terrain itself.

The learned \(\phi(X)\) is then fine-tuned on the downstream task with explicitly acquired labels \((Y_{down})\) as shown in Eq. \ref{eq:01} 
\begin{equation}
\label{eq:01}
    min_\theta = L_{down}(Y_{down},Y^{'}_{down}; \phi(X);\theta)
\end{equation}
    where \(Y^{'}_{down}\) are the predicted labels and \(\theta\) represents the downstream model parameters.

\begin{figure*}
   \centering
    \includegraphics[width=\textwidth, height=10cm]{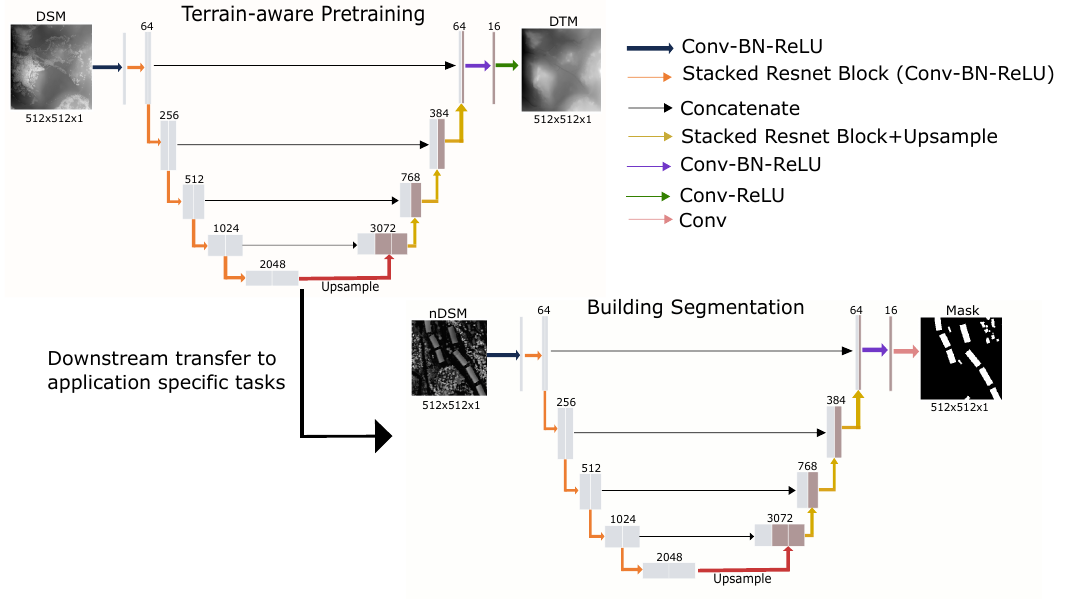}
    \caption{Our approach: We employ U-Net with Resnet-50 encoder for image reconstruction task from LiDAR surface image to terrain image. Through formulating the pretraining as a reconstruction task, all the layers within the encoder and decoder (except for the last task-specific layer) can be utilized in the downstream tasks as shown. \emph{Conv} stands for convolutional operation, \emph{BN} stands for batch normalization.}
    \label{fig:block_diagram}
\end{figure*}
\subsection{Pretraining}
We use U-Net~\cite{ronneberger2015unet} with a ResNet-50~\cite{he2016deep} encoder for the image reconstruction pretext task. The network takes an object raster ($x_o$) as input and predicts the corresponding terrain image ($x_t$) as seen in Figure \ref{fig:block_diagram}. We use an image reconstruction task, as this has the advantage of allowing us to reuse all the layers within the encoder as well as the decoder for our downstream task of segmentation. This is in contrast to with how transfer happens for a classification pretext task \cite{zhao2020self, zhang2018unreasonable}, where only the encoder weights would be useful. The outputs (logits) from the final convolutional layer are used to evaluate the reconstruction loss. We use a combination of two losses to evaluate the reconstruction, the first being a smooth-L1 loss~\cite{girshick2015fast} and the second being LPIPS (learned perceptual image patch similarity) loss~\cite{zhang2018unreasonable}. The smooth-L1 (Eq. \ref{eq:02}) loss converges to either L1 or L2-like loss depending on the hyperparameter $\beta$ chosen as $1.0$ in this experiment. This allows the network to be more forgiving to outliers that L2 loss alone would penalize heavily, while benefiting from L1-like slightly faster convergence.
\begin{equation}
\label{eq:02}
L_{\text{smooth L1}}(y, y') =
\begin{cases}
    \frac{0.5(y - y')^2}{\beta} & \text{if } |y - y'| < \beta \\
    |y - y'| - 0.5\beta & \text{otherwise}
\end{cases}
\end{equation}

The LPIPS loss (Eq. \ref{eq:1}) on the other hand, does not measure pixel-wise similarity between images, but instead measures the perceptual similarity in the target $x_t$ and reconstructed $x_t^{*}$ terrain images. It does so by computing Euclidean distances between the two images in the feature space of pretrained convolutions networks $\mathcal{F}$ such as VGG~\cite{simonyan2014very} or AlexNet~\cite{krizhevsky2012imagenet}.
\begin{equation}
\label{eq:1}
    LPIPS(x_t,x_t^{*}) = \sum_{l \, \in \, \mathcal{F}} w_l \cdot \text{MSE}(\phi_l(x_t), \phi_l(x_t^{*}))
\end{equation}
Here, $\phi_l(\cdot)$ refers to the normalized output of layer $l$ within VGG network $\mathcal{F}$. $w_l$ are the weights for the linear combination of the extracted features and are set to $1$ in our case. Further, while the use of traditional metrics such as SSIM (Structural Similarity Index) \cite{wang2004ssim} and MS-SSIM (Multi-Scale Structural Similarity Index) \cite{wang2003multiscale} are very common as loss functions for a reconstruction task such as this, we found that the negligible advantage obtained was negated by the computation overhead especially when evaluating similarity between the image and its reconstruction at multiple-scales (MS-SSIM). On the other hand, while LPIPS is associated with percepted smoothness, the high perceptual quality is actually a consequence of minimizing the distances between the images at different levels in the feature space. Although perceptual quality is not the primary concern in our case, we trade the computational complexity of MS-SSIM for LPIPS for slightly better reconstructions.

In addition, we use squeeze-and-excitation (SE) attention blocks~\cite{scse} to capture channel-wise dependencies within the feature maps. This can be used to adaptively emphasize some channels while suppressing other less informative ones. SE blocks perform this in two steps via \say{squeeze} and \say{excitation}. The squeeze step uses a global average pooling layer to compress each channel into a scalar descriptor representing the importance of that channel. The channel-wise descriptors are then allowed to interact through fully connected layers to combine and adaptively weight them based on importance. This is the excitation step. The weights can then be applied to original features to rescale them. This has been shown to allow networks the ability to selectively choose among the different feature maps adaptively, which improves segmentation \cite{scse}.

 \section{Evaluation and Results}
 
\begin{figure*}
   \centering
    \includegraphics[width=\textwidth, height=10cm]{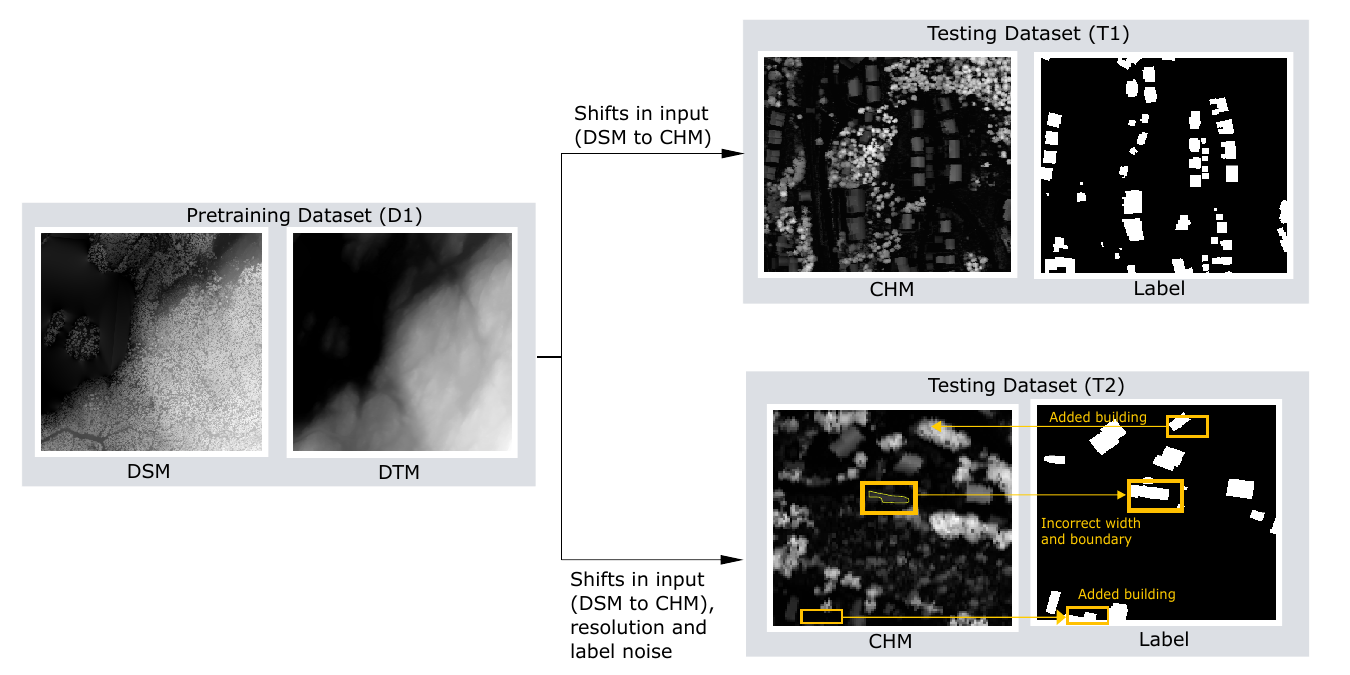}
    \caption{Training dataset (D1) and testing datasets (T1 and T2) : During pretraining the model learns to reconstruct DTM from DSM at 1~m ground resolution. Both testing datasets T1 and T2 input nDSM images for building segmentation, introducing a shift from training data. Further, T2 additionally introduces resolution shift and label noise for testing the generalizability of our approach. Some of the labeling errors can be seen in the figure where added building refers to buildings that are found in the label but are absent in the LiDAR images.}
    \label{fig:dataset}
\end{figure*}

\subsection{Dataset Details}
\subsubsection{Train dataset D1}
Our primary data source consists of DTM and DSM models for Norway at $1m$ ground resolution in the UTM~33N projection. This data is published by the Norwegian Mapping Authority (in Norwegian: Kartverket), and is freely available on their website\footref{hoyde}, under CC~BY~4.0 license. The data is collected as 3D LiDAR point clouds via periodic aerial campaigns, where the maximum offset from the nadir is 20 degrees. Both point clouds and the gridded, reprojected DTM and DSM models are available. In this paper, only the processed DTM and DSM models are used. The subdataset used in this paper is republished under CC~BY~4.0 license ensuring reproducibility.

To acquire labels for the same region, the OpenStreetMap\footnote{\url{https://openstreetmap.org/copyright}}(OSM) database has been used from which building footprints were rasterized to the same grid as the DTM and DSM models. These rasterized labels are published under the Open Database License (ODbL) in accordance with the OSM license requirements.

\subsubsection{Test Datasets}

For testing, we use two different datasets, Norway data T1 acquired in the same way as D1,
and dataset T2 from NORA’s MapAI competition~\cite{jyhnemapai}. Compared to the training dataset, T2 reflects acquisition and distortion shifts including but not limited to varying quality and different types of label noise (incomplete and incorrect labels, shown in Figures \ref{fig:dataset} and \ref{fig:errors}). Note that the MapAI dataset's exact resolution is unknown, and the data is not georeferenced. In this work, we use only a subset of the total challenge dataset (LiDAR images and mask for task 2 in the challenge, the accompanying spectral images have not been used as our work focuses on building detection from LiDAR data only) as T2. The train and test dataset curated for this work can be accessed \href{https://dataverse.no/}{\textit{here}}.

 This section discusses the evaluation protocol and results from the terrain-aware pretraining. We use two metrics as indicators of performance given by IoU (Intersection over Union) and bIoU (boundary IoU). IoU is defined as the ratio of the intersection of two sets to their union (Eq. \ref{eq:2}). In segmentation, the two sets are the sets of pixels that are marked as the predicted object, $Pred$, and the ground truth, $GT$, respectively.
\begin{equation}
    \label{eq:2}
    IoU = \frac{|Pred \cap GT|}{|Pred \cup GT|}
\end{equation}
bIoU extends the metric to evaluate the boundaries (external edge of the buildings) more specifically and is thus IoU over masks considering a specified pixel thickness $d$ along the boundaries.

\subsection{Details on Pretraining and Training}
 The pretraining dataset consists of 16323 images for training and 1813 for validation with a patch size of 512x512 pixels. We use the Adam optimizer with an initial learning rate of $1 \times {10^{-6}}$ and weight decay of$1 \times {10^{-8}}$  at a batch size of 8. The learning rate is reduced by a factor of $0.1$ if the validation loss does not improve over 10 epochs. We use gradient clipping, where the gradients are capped at the maximum gradient norm value of 1. The pretraining converged around 300 epochs at which point the IoU and bIoU on the test split were 0.933 and 0.858 respectively. The pretrained features are evaluated for performance and generalizability on test datasets T1 and T2, where T1 tests performance under varying few-shot schemes and T2 captures the generalizability of the proposed approach under domain shifts. For evaluation of the target task of building segmentation, the pretrained U-Net is modified from the terrain reconstruction task to perform segmentation by adding a segmentation head (convolution block followed by ReLU activation). For finetuning, we use RMS prop optimizer (momentum of 0.999 and weight decay $1 \times {10^{-8}}$ ), initial learning rate $1 \times {10^{-6}}$  which decays by a factor of 10 every 15 epochs if the loss does not decrease. Weighted cross entropy loss and DICE loss are used to evaluate segmentation performance~\cite{sudre2017DICE}. The model used for pretraining has approximately 24 million parameters and processed approximately 6 images/second on a TWIN-TITAN RTX (2 GPUs). The downstream segmentation model is also of similar complexity with approximately 33 million parameters and processes approximately 10 images/second both during training and inference.

\subsection{Segmentation Results}

One of the focus areas in this work is label-efficiency, whereby if the pretraining is effective, good performance can be achieved on the downstream task with relatively few labels. We test this by performing segmentation on T1 using nDSM as input instead of DSM. The use of nDSM instead of DSM (as in pretraining) introduces a slight shift in train and test data, making the task a little more challenging.  In Table I, we present \textbf{ablations} focusing on label percentages, thereby assessing performance under conditions of low-label availability.

Table \ref{tab:100_norway} shows the performance of building segmentation on T1 with  different label fractions under full fine-tuning such that all parameters are tuned during this evaluation. Remarkably, with 1\% of the labels (equivalent to merely 25 labels), our method achieves similar performance (slightly higher by a margin of 0.03 IoU and similarly for bIoU) compared to pretraining with 1.2 million images and 1,000 different categories. This demonstrates the advantage of learning features from freely available unlabelled data in the domain of remote sensing by curating specific tasks that enable learning features specialized to downstream tasks, to an extent where merely 25 examples are sufficient for target task generalization.

The second and third rows in table \ref{tab:100_norway} show the segmentation performance on 10\% (252 examples) and 100\% (2500 examples) label fractions. The advantage of pretext pretraining is more apparent in the few-shot schemes (1\% and 10\%), and as the labels increase, our method closes the gap with ImageNet pretrained features.

\begin{figure}[!htbp]
   \centering
    \includegraphics[width=2.8in, height=9in]{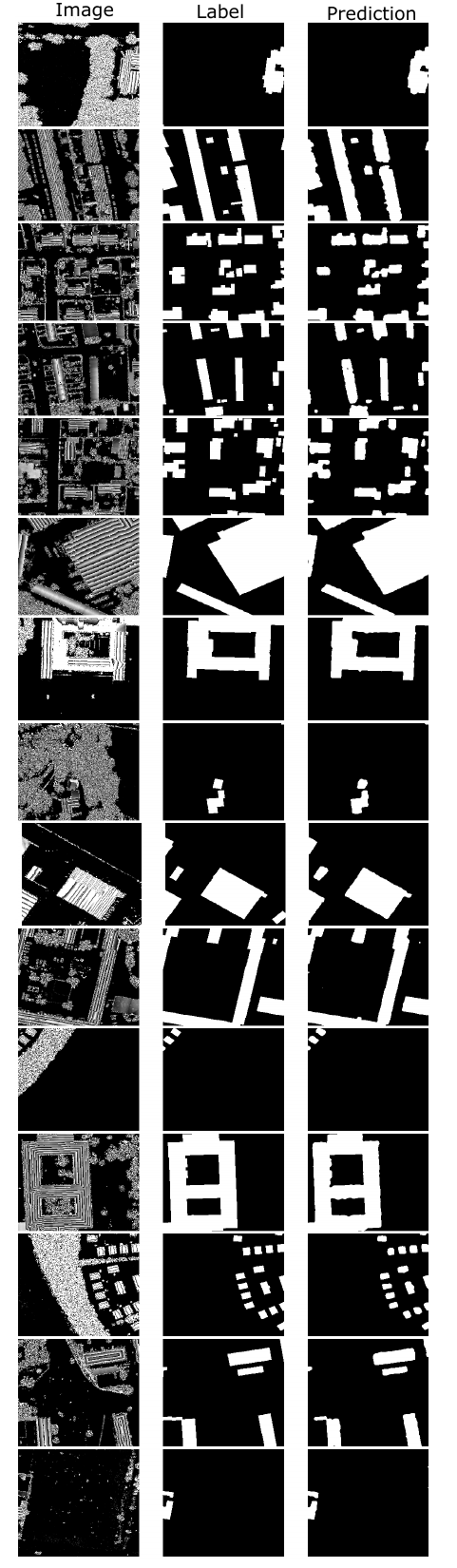}
    \caption{Predicted segmentation masks from our approach on T2 (MapAI) dataset.}
    \label{fig:result}
\end{figure}

\begin{table}[!htbp]
    \renewcommand*{\arraystretch}{1.4}
    \resizebox{0.49\textwidth}{!}{
        \centering
        \begin{tabular}{ l l l l l l l}
            \hline
           &              \multicolumn{3}{c}{Validation} & \multicolumn{3}{c}{Test} \\ \hline
                        \hline
             Approach & IoU            & bIoU    & Score & IoU                 & bIoU  & Score \\ \hline
            \multicolumn{7}{c}{ 1\% (25 labels) on T1 data}\\
            \hline
             ImageNet   & 0.742          & 0.640    & 0.691 & 0.772               & 0.641 & 0.707 \\ 
              Ours                   & \textbf{0.775} & 0.551 & 0.663   & \textbf{0.794} & 0.507 & 0.650 \\ \hline
            \multicolumn{7}{c}{10 \% (252 labels) labels on T1 data} \\
            \hline
              ImageNet   & 0.851    & 0.777   & 0.814 & 0.848              & 0.748          & 0.798 \\ 
              Ours                   & 0.831    & 0.731   & 0.781 & \textbf{0.860} & \textbf{0.769} & \textbf{0.842} \\ \hline
            \multicolumn{7}{c}{ 100\% (2500 labels) labels on T1 data}\\
            \hline
             ImageNet   & \textbf{0.901}  & 0.859 & 0.880  & \textbf{0.9} & 0.842 & 0.871 \\ 
             Ours                   & \textbf{0.896}         & 0.834 & 0.857 & \textbf{0.896}       & 0.835 & 0.866 \\ \hline
        \end{tabular}
    }
    \caption{Segmentation performance with full fine tuning with different percent of ground truth labels on T1 data compared with our approach against ImageNet pretraining.}
    \label{tab:100_norway}
\end{table}

\begin{figure}[!htbp]
   \centering
    \includegraphics[width=3.2in, height=7.5in]{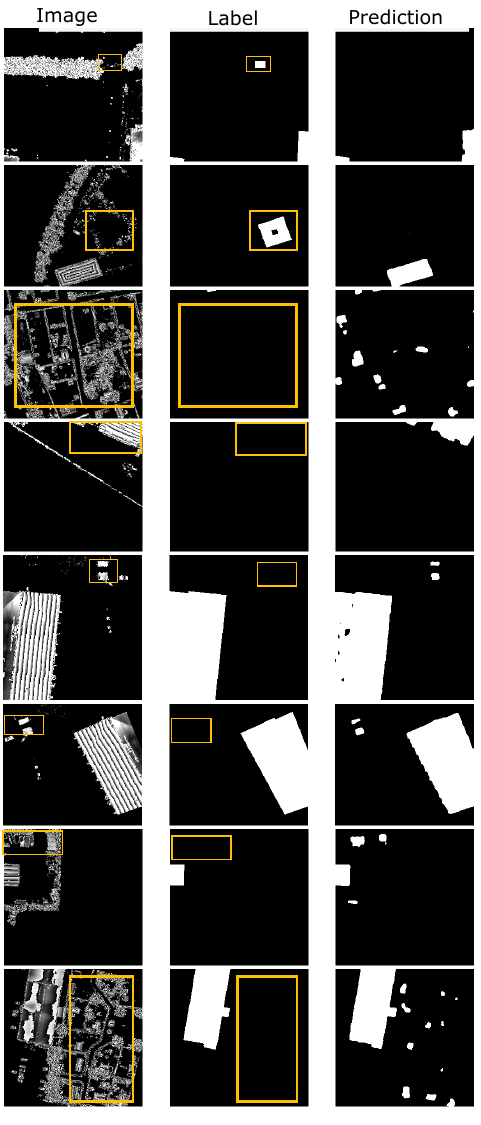}
    \caption{Images from T2 with labeling inaccuracies and our approach's segmentation masks. As seen in column two, the labels do not accurately reflect the buildings in the images. These inaccuracies are highlighted with orange bounding boxes and signify either absent buildings or buildings marked in the labels but absent in the images. However, despite these errors, our method exhibits robustness in detecting these missing or incorrectly labeled buildings, as observed in the third column. We assume that in the presence of such labels, our reported performance is slightly understated due to incorrect penalization for several of the correct predictions.}
    \label{fig:errors}
\end{figure}

\subsection*{Generalizability Evaluation}
\begin{figure}[!htbp]
    \centering
    \includegraphics[width=3.486in]{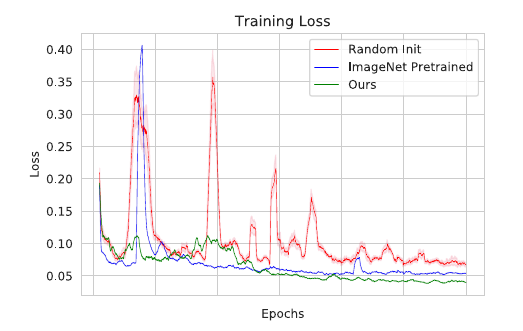}
    \caption{Training loss of building segmentation on T2, with three different network initializations. \emph{Random Init} refers to initializing the network from random weights, \emph{ImageNet pretraining} refers to initializing the network with weights from training on ImageNet classification. \emph{Ours} refers to initializing the network with weights pretrained on terrain prediction task. }
    \label{fig:training}
\end{figure}

As discussed in Section \ref{sec:intro}, the other challenge addressed in this work is generalization. In remote sensing, this implies countering shifts that can occur due to different input resolutions, temporal variabilities, labeling variations, as well as errors. In this section, we evaluate the generalizability of pretrained features on T2~\cite{jyhnemapai} which captures a significant distribution shift compared to the training dataset (refer to figure \ref{fig:scales}).
\begin{figure*}[!htbp]
    \centering
    \includegraphics[width=\textwidth]{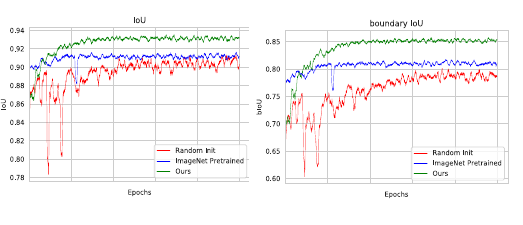}
    \caption{Training performance metrics of building segmentation on T2, with  network initializations as described in Figure \ref{fig:training}.}
    \label{fig:iou}
\end{figure*}
Further, we also test for label efficiency and computational efficiency of our approach. To check for label efficiency we compare with U-Net pretrained on ImageNet classification. To evaluate the computational efficiency and quality of features, we compare against two other approaches that perform building segmentation on T2~\cite{li2023buildseg, borgersen2022mapai}. Further, we also perform an ablation against U-Net with random initialization to evaluate the benefit obtained from pretraining on pretext task in terms of downstream convergence speed and ease of optimization (Figures \ref{fig:training} and \ref{fig:iou}). Our method performs better than ImageNet-based U-Net by a large margin (Table \ref{tab:mapai}; 0.323 in IoU and 0.367 bIoU). Some labels in T2 correspond to real-time labeling errors where buildings in aerial images do not correspond to the available ground truth masks (Figure \ref{fig:errors}). Further, the masks are generated not from aerial images directly, but from a DTM. As a result, building displacement errors can be expected in the dataset. The low performance of ImageNet pretrained model can be attributed to such label noise in addition to domain shift arising from translating natural image features to LiDAR data. In contrast with~\cite{borgersen2022mapai}, which uses both spectral and LiDAR data for building segmentation, we see an improvement by a margin of 0.155 in IoU. This confirms the initial idea that it can be advantageous to use LiDAR data independently, without augmenting it with other spectral information for segmentation problems.

We further show how our pretraining strategy along with a simplistic U-Net allows us to achieve better or comparable performance to significantly more complex architectures~\cite{li2023buildseg}. Our terrain-aware U-Net model performs better than the transformer-based segmentation model (Segformer)~\cite{xie2021segformer}, which leverages a self-attention mechanism to capture long-range dependencies between pixels, which makes it suitable for capturing contextual information in semantic segmentation tasks. This shows that careful pretraining can help achieve computational cost and power efficiency in the long run, allowing much smaller models to achieve similar or better performance. Furthermore, features from the terrain-aware pretraining task can be leveraged across a variety of downstream tasks, allowing feature reuse and enhancing label efficiency even further.

\begin{table}[!htbp]
    \renewcommand*{\arraystretch}{2}
    \resizebox{0.489\textwidth}{!}{
    \tiny
        \centering
        \begin{tabular}{l l l l l}
            \hline
            \textbf{Model}                     & \textbf{IoU}   & \textbf{bIoU}  & \textbf{Score} \\
            \hline
            \hline
            U-Net (Random)                     & 0.470          & 0.261          & 0.366          \\
            \hline
            U-Net (ImageNet pretrained)       & 0.521          & 0.326          & 0.423          \\
            U-Net~\cite{li2023buildseg}        & 0.761          & 0.582          & 0.672          \\
            U-Net~\cite{borgersen2022mapai}    & 0.689          & 0.562          & 0.625          \\
            ConvNext~\cite{li2023buildseg}     & 0.784          & 0.610          & 0.697          \\
            SegFormer-B0~\cite{li2023buildseg} & 0.763          & 0.590          & 0.676          \\
            SegFormer-B4~\cite{li2023buildseg} & 0.784          & 0.611          & 0.698          \\
            SegFormer-B5~\cite{li2023buildseg} & 0.790          & 0.618          & 0.704          \\
            \hline
            U-Net (Ours)                       & \textbf{0.844} & \textbf{0.693} & \textbf{0.768} \\
            \hline
        \end{tabular}
    }
    \caption{Evaluation on Map AI challenge data with full fine tuning}
    \label{tab:mapai}
\end{table}

We further compare the three network initializations (random, ImagetNet, and our terrain initializations) to study their impact on training stability and speed. On T2 (Figure \ref{fig:training}), our pretraining strategy exhibits greater training stability at the earlier optimization stages. This suggests that our pretraining allows starting from a more optimal point on the loss landscape compared to others.
Further, we observe faster convergence, especially when compared to training from random weights. This leads to a more efficient utilization of computational power and resources. The same effect is observed for the performance metrics IoU and bIoU (Figure \ref{fig:iou}).

\section{Conclusion}
Motivated by the ever-increasing demand for accurate and up-to-date urban planning and development information, in this work we propose leveraging self-supervised learning for efficient building footprint extraction from LiDAR data. Our terrain-aware self-supervised learning task shows improvement with regard to both generalization and label efficiency. Further, we show that pretraining on completely unlabeled data from LiDAR enables the extraction of domain-related features that compete and improve over supervised approaches. In the few-shot evaluation, with as few as 25 labeled examples, our approach rivals or surpasses the traditional ImageNet pretrained models (Tables \ref{tab:100_norway} to \ref{tab:mapai}), bigger architectures (Table \ref{tab:mapai}) and custom-tailored augmentation strategies \cite{borgersen2022mapai} on building segmentation. Further, as opposed to approaches combining data from different sensors, in this work we use LiDAR data independently, indicating the usefulness of depth information inherent in LiDAR for building segmentation. The data used in this article can be accessed at \url{https://dataverse.no/} (DOI: 10.18710/HSMJLL).


\section*{Acknowledgment}
This paper was funded by the Research Council of Norway under the project title ATELIER-EO: Automated machine learning framework tailored to Earth Observation [project number : 336990].


\bibliographystyle{IEEEtran}
\bibliography{bibtex/bib/IEEEexample}


\end{document}